\documentclass[runningheads]{llncs}
\usepackage{graphicx}
\usepackage{latexsym}
\usepackage{booktabs}
\usepackage{comment}
\usepackage{multirow}
\usepackage{subcaption}

\usepackage{appendix}
\usepackage{amsmath}
\usepackage{bm}
\usepackage{tabularx}
\usepackage{colortbl}
\usepackage{tikz}
\usepackage{stfloats} 
\usepackage{url}

\usepackage{microtype}

\DeclareMathOperator*{\argmax}{arg\,max}

\usepackage{xspace}
\newcommand{\eg}{e.g.\@\xspace}
\newcommand{\ie}{i.e.\@\xspace}
\newcommand{\gpttwo}{\mbox{GPT-2}\xspace}
\newcommand{\gptthree}{\mbox{GPT-3}\xspace}
\newcommand{\structshot}{\textsc{StructShot}\xspace}

\definecolor{sandybeach}{RGB}{250,245,230}
\definecolor{darkgreen}{RGB}{20,160,20}
\definecolor{powdergray}{RGB}{235,235,235}
\definecolor{bestgray}{RGB}{220,220,220}

\begin{document}
\title{TOKEN is a MASK: Few-shot Named Entity Recognition with Pre-trained Language Models}
\titlerunning{TOKEN is a MASK}
%
\author{Ali Davody\orcidID{0000-0003-2687-4577}\inst{1,2} \and
David Ifeoluwa Adelani\orcidID{0000-0002-0193-2083
} \inst{1} \and
Thomas Kleinbauer\inst{1} \and Dietrich Klakow\inst{1}}
%
\authorrunning{A. Davody et al.}
%
\institute{Spoken Language Systems Group, \\ Saarland Informatics Campus, Saarland University, Saarbrücken, Germany.
\and
Testifi.io, München, Germany\\
\email{\{adavody,didelani,thomas.kleinbauer,dietrich.klakow\}@lsv.uni-saarland.de}\\
}

\maketitle              
\begin{abstract}
Transferring knowledge from one domain to another is of practical importance for many tasks in natural language processing, especially when the amount of available data in the target domain is limited. In this work, we propose a novel few-shot approach to domain adaptation in the context of Named Entity Recognition (NER). We propose a two-step approach consisting of a variable base module and a template module that leverages the knowledge captured in pre-trained language models with the help of simple descriptive patterns. Our approach is simple yet versatile, and can be applied in few-shot and zero-shot settings.
Evaluating our lightweight approach  across a number of different datasets shows that it can boost the performance of  state-of-the-art baselines by $2-5\%$ F1-score.
\end{abstract}
\section{Introduction}
\label{sec:intro}



Transfer learning has received increased attention in recent years because it provides an approach to a common problem for many realistic Natural Language Processing (NLP) tasks: the shortage of high-quality, annotated training data. While different implementations exist, the basic tenet is to utilize available data in a source domain to help 
training
a classifier for a low-resource target domain.

An
interesting new direction of research leverages the world knowledge captured by pre-trained language models (PLMs) with cloze-style natural language prompts for few-shot classification (\eg \cite{brown-mann+-2020-lm-few-shot,
schick-schutze-2021-cloze}) and regression \cite{gao-fisch-chen-2020-lm-few-shot}. These approaches are attractive because they require little to no training, making them especially suitable for low-resource settings.

In this paper, we contribute to this research area by introducing a novel cloze-style approach to Named Entity Recognition (NER), an important task which has previously not been addressed via cloze-style prompts. In its classical setting, \ie recognizing a small number of entity types in newspaper texts, NER 
achieves
state-of-the-art F1 scores of $\sim{}95\%$ \cite{yamada-etal-2020-luke}. This is not necessarily the case, however, for more specialized domains where data is more scarce and annotations cannot easily be provided because they may require expert knowledge, such as \eg, for biomedical texts. With the approach presented here, the expertise of highly trained specialists can be utilized in a different way, by providing \textit{representative words} for the named entity types, rather than having to annotate corpus data.

The main appeal of our method lies in its simplicity, as applying it to a new domain requires very little effort and technical expertise. 
Our contribution 
is three-fold: (1) we introduce a new method for Named Entity Recognition (NER) with a focus on simplicity; (2) our technique is scalable down to zero-shot in which case \emph{no training} is required on top of the PLM; (3) we show how a hybrid combination of our method with a standard classifier based on a simple threshold outperforms both of the individual classifiers (Section~\ref{sec:method}).

The effectiveness of our method is demonstrated by a thorough evaluation comparing different variants of the approach across a number of different data sets (Section~\ref{sec:experiments}). For reproducibility, we release our code on Github\footnote{\url{https://github.com/uds-lsv/TOKEN-is-a-MASK}}



\section{Related Work}
\label{sec:related-work}

Named entity recognition is a well-studied task in NLP, and is usually approached as a sequence-labeling problem where pre-trained language models such as ELMO~\cite{peters-etal-2018-deep}, BERT~\cite{devlin-etal-2019-bert},  RoBERTa~\cite{Liu2019RoBERTaAR} and LUKE \cite{yamada-etal-2020-luke} have brought significant improvements in recent years. 
All these methods are based on supervised learning but they do not generalize to new domains in zero and few-shot settings. 

Meta-learning or \textit{learning to learn}~\cite{Ravi2017OptimizationAA,Finn2017ModelAgnosticMF,NIPS2017_snell} is a popular approach to few-shot learning. 
In the context of few-shot NER, most applications of meta-learning make use
of Prototypical Networks~\cite{Fritzler_ner,hou-etal-2020-shot,Huang2020FewShotNE} or Model-Agnostic Meta-Learning (MAML)~\cite{krone-etal-2020-learning}. These approaches require training on diverse domains or datasets to generalize to new domains.   

Pre-trained language models have shown impressive potential in learning many NLP tasks without training data~\cite{petroni2019language,radford2019language}. 
\cite{schick-schutze-2021-cloze} proposed using a cloze-style question to enable masked LMs in few-shot settings to perform text classification and natural inference tasks with better performance than \gptthree~\cite{brown-mann+-2020-lm-few-shot}. 
As creating cloze-style questions is time consuming, there are some attempts to automate this process. 
\cite{gao2020making} makes use of the T5 model~\cite{Raffel2020ExploringTL} to generate appropriate template by filling a [MASK] phrase similar to how T5 was trained.  Shin et al. (2020)~\cite{shin-etal-2020-autoprompt} use a template that combines the original sentence to classify with some trigger tokens and a [MASK] token that is related to the label name. The trigger tokens are learned using gradient-based search strategy proposed in \cite{Wallace2019UniversalAT}. 
In this paper, we extend this PLM prompt technique to named entity recognition. 

\section{Method}
\label{sec:method}

Our approach consists of two parts. We first describe the \emph{base method}
that can be used as a stand-alone, zero- or few-shot classifier (Section~\ref{sec:base-method}). In Section~\ref{sec:hybrid-method}, we then lay out how a simple ensemble method can combine the base method with another classifier to potentially improve over the individual performance of both. We call this setup the \emph{hybrid method}.

\subsection{Zero-Shot Base Method}
\label{sec:base-method}

The base method for classifying NEs in a sentence consists of two steps, \textit{detecting} candidate words and \textit{querying} a PLM for the NE class of each candidate word. For example, let $s =$ \textit{I will visit Munich next week} be the sentence to label. As is typical, most words in $s$ do not denote a named entity, only \textit{Munich} does (\texttt{LOC}). 
We construct a cloze-style query to the PLM from the original sentence and a template, in which the candidate word has been inserted:

\vspace{2mm}
\begin{center}
  \begin{minipage}{.9\columnwidth}
    \textit{\underline{I would like to visit Munich next week.} \underline{Munich} is a [MASK].}
  \end{minipage}
\end{center}
\vspace{2mm}

The first part of the prompt 
serves as the context for the second part of the prompt, a template of a predefined form \eg \texttt{[TOKEN] is a [MASK]}. 
The \texttt{[TOKEN]} is replaced with the term to label. For auto-regressive LMs like GPT-2, \texttt{[MASK]} is empty, the next word is predicted given  a context that ends with ``TOKEN is a ''. 

\medskip
The prediction of the PLM for the [MASK] is a probability distribution $P(w|s)$ over the tokens $w$ in the vocabulary $\mathcal{V}$. Intuitively, since NE labels themselves  often are descriptive words (\eg \textit{location, person}, etc.) contained in $\mathcal{V}$, the answer to the query could be found by observing $P(label|s)$ for all labels and selecting the one with the highest probability. 
We found this 
approach not to perform well, possibly because NE labels tend to describe abstract higher-level concepts that are not commonly used in sentences realized by our template. 


However, by associating with each entity type a list of words representative of that type, we reach competitive performance to state-of-the-art approaches (see Section \ref{sec:experiments}). As an example, table \ref{tab:words} shows representative words for the five named entity classes \texttt{Location}, \texttt{Person},  \texttt{Organization}, \texttt{Ordinal} and \texttt{Date}.

More formally, let $\mathcal{L}$ be the set of all labels for the NER classification task. We provide a list of representative words $\mathcal{W}_l$ for each label $l$. Denoting the output of the PLM by $P(\cdot|s+T(w))$ where $s$ is the original sentence, $T(w)$ is the prompt for token $w$, and $+$ stands for string concatenation, we assign label $l_w$ to the token $w$ by:

\vspace{-5mm}
\begin{equation}
    l_w = \argmax_{l} P(v \in \mathcal{W}_l|s+T(w)).
\end{equation}

\begin{figure}[t]
\centering
\includegraphics{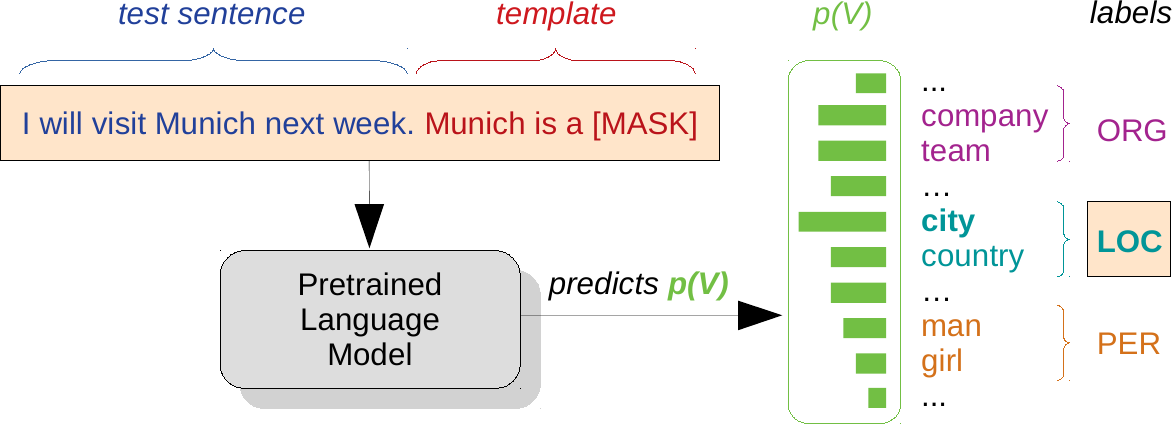}
\vspace{-2mm}
    \caption{Overview of the proposed template method: The PLM prompt  consists of the sentence and the template instantiated with the word to label. The PLM predicts the probability of [MASK] for all representative words, and the label associated with the most probable word is returned.}
    \label{fig:method-overview}
\end{figure}

\noindent
For the example above ($s=$ \textit{I will visit Munich next week}; $T(\textit{Munich})=$ \textit{Munich is a [MASK]}), the top-5 predictions using the \textit{BERT-large model} are: \texttt{city, success, democracy, capital, dream}. The largest probability ($0.43$) among all words is assigned to \texttt{city} which is among the representative words for label \texttt{LOC}. Thus, \textit{Munich} is labeled as a location. A graphical depiction of the full method is shown in Figure \ref{fig:method-overview}.


\medskip

\noindent
The outlined approach raises three design questions which we address in turn:
\begin{enumerate}
\item Given the input sentence, how to detect the candidate tokens to label?
\item What constitutes a good template for the second half of the prompt?
\item Which words to include in the list of representative words for each entity label?
\end{enumerate}
A fourth decision to make is the choice of pre-trained language model. Our results comparing four state-of-the-art PLMs are listed in section \ref{sec:comparing-lms}, but we shall first attend to the three questions listed above.

\begin{table}[tb]
 \centering
 \footnotesize
  \begin{tabular}{ll}
    \toprule
    \textbf{Entity Type} & \textbf{Representative Words} \\
    \midrule
    LOC & location, city, country, region, area, province, state, town\\
    PER & person, man, woman, boy, girl, human, someone, kid\\
    ORG & organization, community, 
    department, association, company, 
    team \\
    ORDINAL & number, digit, count, third, second\\
    DATE & date, day, month, time, year\\
    \bottomrule
  \end{tabular}
  \vspace{1mm}
  \caption{Examples of representative word lists for entity types
  \texttt{location} (LOC), \texttt{person} (PER),  \texttt{organization} (ORG),  \texttt{Ordinal} (ORDINAL), and \texttt{date} (DATE).}
  \label{tab:words}
  \vspace{-2mm}
\end{table}

\paragraph{Identifying Named Entities.} 

In sequence labeling tasks, it is common that the identification of token boundaries and the actual label of the token is performed jointly as both decisions are often informed by the same features. Here, we perform these steps separately. This is possible because named entities are usually realized as proper nouns (although some NE schemes also include entities such as numbers, date, and time expressions), licensing the use of a task-independent part-of-speech tagger to identify candidate tokens in the input sentence before labeling it with the template method. Part-of-speech (POS) \footnote{We make use of Spacy POS tagger  \url{https://spacy.io/usage/linguistic-features}} taggers can identify occurrences of proper nouns (or numerals, ordinals) with a high degree of accuracy. However, entity boundaries in case of multi-word expressions are usually not labeled explicitly. A phrase structure parser could be employed to determine the boundaries of multi-word named entities, but we found simply treating consecutive proper nouns as a single entity to be a reasonable heuristic in our experiments. 

\paragraph{Template Selection.} In order to gain insights into what constitutes a good template we have experimented with a number of variants. Overall, we found that most templates performed similarly, except where we intentionally tested the limits of the format. These experiments are detailed in Section~\ref{sec:template-choice}.
Generally speaking, a template should be grammatical and natural. Simple variants work better than overly complicated templates. The simple \mbox{copula} \texttt{[TOKEN] is a [MASK]} is a good recommendation as it performed well in all of our tested conditions. 

\paragraph{Representative Words.} If some examples of the target domain are available at training time, the representative word lists can be derived from predictions over this data. That is, the representative words for each entity type can be taken as the most probable mask fillers given the respective prompts. Alternatively, in a zero-shot setting or where otherwise appropriate, a set of words can be provided by a domain expert. For the experiments in the next section, this is the strategy we followed. Of course, it is possible to combine both methods, with an expert choosing a suitable subset from words with the highest probability in the automatic setting. Additionally, we can use static word embeddings like GloVe to extract a list of representative for each category based on word similarity.

\subsection{Few-shot Hybrid Method}
\label{sec:hybrid-method}

The method described thus far can be used for zero-shot classification as it does not require any training. In a few-shot setting, however, we can improve the performance of the system by fine-tuning the PLM using the labeled data in the target domain. 

A further performance gain can be made by combining our method with a standard supervised classifier in a simple two-fold ensemble. Based on a selection threshold $p_h$, we label the token according to the prediction of the base method if the probability of the predicted label is higher than the threshold $p_h$:
\begin{equation}
    \max_{l} P(v \in \mathcal{W}_l|s+T(w)) > p_h,
\end{equation}
otherwise we relay the output of the supervised classifier. The threshold $p_h$ is a hyper-parameter that can be tuned on the training examples from the target domain.


\section{Data}
\vspace{-2mm}
\label{sec:data}

We consider three popular NER datasets 
(CoNLL03, OntoNotes~5.0 and i2b2) 
from different domains and with different label sets. 

The \textbf{CoNLL03} data~\cite{tjong-kim-sang-de-meulder-2003-introduction} consists of documents from the English news domain that have been annotated with four named entity types: personal name (PER), organization (ORG), location (LOC), and miscellaneous entity (MISC).

The \textbf{OntoNotes 5.0} dataset~\cite{pradhan-etal-2013-towards} consist of NER in three languages (English, Chinese and Arabic). In this work we make use of the English dataset with six annotated domains: broadcast news (BN), broadcast conversation (BC),  magazines (MZ), telephone conversation (TC), and web data (WB). The annotation scheme of OntoNotes~5.0 distinguishes between $18$ entity types, making it a more challenging task than CoNLL03. 


The third dataset, \textbf{i2b2 2014}~\cite{stubbs_i2b2}, is a BioNLP dataset commonly used for de-identification tasks. The dataset contains $25$ entity types from seven Private Health Information (PHI) categories: Name, Profession, Location, Age, Date, Contact, and Ids. 

In our few-shot experiments, $1$, $5$, or $100$ data points are sampled per label from each training set, depending on the setting. In the zero-shot case, no training is used.

\section{Experiments}
\vspace{-1mm}
\label{sec:experiments}

 \begin{table}[t]
 \footnotesize
 \centering
    \scalebox{0.9}{
  \begin{tabular}{l@{\hspace{1ex}}cccc}
    \toprule
    \textbf{} & \textbf{BERT} &\
    \textbf{RoBERTa}  & \textbf{\gpttwo}    & \textbf{XLNet}  \\
    \midrule
    LOC & 69\% & 65\% & 42\% &   58\%\\
    PER & 80\% & 73\% & 45\% &  57\% \\
    ORG & 42\% & 43\% & 13\% &  34\%  \\
    \midrule
    Micro-Avg. & 60\% & 59\% & 36\% &  49\% \\
    \bottomrule
  \end{tabular}
  }
  \vspace{2mm}
  \caption{F1-scores for four PLMs on the CoNLL03 dataset in the zero-shot setting.}
  \vspace{-5mm}
  \label{tab:lm_result}
\end{table}

\subsection{Comparing Language Models}
\label{sec:comparing-lms}

We study the role of the choice of PLM in our approach by comparing four state-of-the-art pre-trained language models: BERT~\cite{devlin-etal-2019-bert}, RoBERTa~\cite{Liu2019RoBERTaAR}, \gpttwo~\cite{radford2019language} and XLNET~\cite{NEURIPS2019_xlnet}. BERT and RoBERTa are trained to predict masked tokens that are randomly corrupted in the training sentence. \gpttwo is based on autoregressive language modeling, but it is less efficient predicting masked tokens. XLNET attempts to address the limitations of BERT for next-word prediction (i.e autoregressive LM)  while retaining good performance on natural language understanding tasks. 

Table \ref{tab:lm_result} compares the four PLMs on the CoNLL03 dataset for the zero-shot setting as described above. In all of these experiments, the template \texttt{[TOKEN] is a [MASK]} and the representative word lists from Table~\ref{tab:words} were used.
We observe that BERT and RoBERTa outperform the other two LMs. Interestingly, \gpttwo does not perform well in this setting with an average F1 score of just $36\%$.


\subsection{Choice of Template}
\label{sec:template-choice}

\begin{table}[t]
\centering
 \scalebox{0.9}{
\centering
\begin{tabular}{c>{\raggedright\arraybackslash}l}
\toprule
\textbf{ID} & \textbf{Template} 
\\
\midrule
$T_1$ & [TOKEN] is a [MASK].
\\
$T_2$ & [TOKEN] was a [MASK]. 
\\
$T_3$ & [TOKEN] would be a [MASK].
\\
$T_4$ & [TOKEN] a [MASK].
\\
$T_5$ & [TOKEN] [MASK].
\\
$T_6$ & [TOKEN] is an example of a [MASK].
\\
$T_7$ & [TOKEN]  is an instance of a [MASK].\\
$T_8$ & [TOKEN] denotes a [MASK].\\
$T_9$ & [TOKEN] is well-known to be a [MASK].\\
$T_{10}$ & Many people consider [TOKEN] to be a [MASK].
\\
$T_{11}$ & [TOKEN] is a common [MASK] known to many people. 
\\
$T_{12}$ & There are many [MASK]s but [TOKEN] stands out nevertheless.
\\
$T_{13}$ & A [MASK] like [TOKEN] is often mentioned in conversations.
\\
$T_{14}$ & A [MASK] like [TOKEN]. 
\\
$T_{15}$ & This [MASK], [TOKEN], is worth discussing.
\\
\bottomrule
\end{tabular}
}
\vspace{1mm}
\caption{List of templates used in the comparative experiments. Aspects considered include tense, mood, expression length and complexity, verb choice, grammaticality, and TOKEN/MASK order, in different forms.}
\label{tab:template-list}
\end{table}

A second variable in the setup is the choice of the template used in the prompt. We compare 15 different templates in the zero-shot setting, listed in Table~\ref{tab:template-list}. We examined these templates on the CoNLL03 and OntoNotes datasets using \texttt{location}, \texttt{person} and  \texttt{organization} as entity classes. 
Tables~\ref{tab:template-results-CoNLL} and \ref{tab:template-results-ontonotes} present the results of this experiment.
%

$T_1$  is a straight-forward default and copula template that directly identifies the token with the mask. $T_2$ template differs from $T_1$ only in the verb. We added it to study the influence of tense on the prediction quality. In our experiments, it lead to slightly worse results, but it might be useful \eg for historical texts. $T_3$ is similar to the above but uses a modal auxiliary. We further experimented with stripping down a template to the minimum useful form in $T_4$. This template is ungrammatical, however, and does not perform well. $T_5$ is the smallest template possible modulo word order. It is the worst performing template in our experiments, especially failing to predict the \texttt{PER} class.
The next four templates $T_6-T_9$ are further variations of $T_1$ that replace the verb \textit{to be} with longer constructions.

In all previous templates, the token to label appears as the first word. In $T_{10}$, we test whether a longer left-hand side context is beneficial to the PLM prediction. With $T_{11}$, we test the effect of extending the right-hand context. It does not produce the same performance gain as $T_{10}$, though.  $T_{12}$  extends both the left-hand and the right-hand side context simultaneously, but also presents token and mask in a contrasting relation. It seems that the language model has more difficulties associating the mask and the token with each other in this template, as the performance drops considerably for CoNLL03.  $T_{13}$ also reverses the order of mask and token but outperforms $T_{12}$ template in the case of CoNLL03.

Are the additional filler words responsible for the good performance of $T_{13}$, or is it the way the relation between mask and token are expressed using the word \textit{like}? $T_{14}$ reduced template suggests the latter, as it performs even slightly better than $T_{13}$. Finally $T_{15}$ is similar in spirit to ${T_5}$ in that it tests whether proximity of mask and token are important, only with the order of the two reversed, and some context words added. It performs better than ${T_5}$ but not en par with most other templates.

The key message of these experiments is that our approach is robust against the details of the template format as long as it is not too artificial. Indeed, we do not observe a high variation in the performance of the model when using different natural sounding templates.

\begin{table*}[tbp]
  \centering
  \footnotesize
  \scalebox{0.90}{
    \centering
    \begin{tabular}{p{16mm}ccccccccccccccc}
      \toprule
      \textbf{} & \textbf{$T_1$} &\textbf{$T_2$}  & \textbf{$T_3$}
      & \textbf{$T_4$} &\textbf{$T_5$} &\textbf{$T_6$}  &\textbf{$T_7$} &\textbf{$T_8$} & \textbf{$T_9$}& \textbf{$T_{10}$} & \textbf{$T_{11}$}
      & \textbf{$T_{12}$}  & \textbf{$T_{13}$} & \textbf{$T_{14}$} & \textbf{$T_{15}$}\\
      \midrule
      LOC & 69\% & 60\% & 62\% & 52\% & 46\% & 66\% & 63\% & 60\% & 67\% & 69\% & 61\% & 57\% & 73\% & 70\% & 60\%\\
      PER & 80\% & 72\% & 73\% & 65\% & 7\% & 81\% & 82\% & 71\% & 76\% & 82\% & 83\% & 15\% & 76\% & 81\% & 57\%\\
      ORG & 42\% & 48\% & 51\% & 35\% & 35\% & 44\% & 39\% & 41\% & 47\% & 44\% & 41\% & 36\% & 48\% & 54\% & 34\%\\
      \midrule
      Micro-Avg. & 60\% & 57\% & 57\% & 47\% & 31\% & 59\% & 57\% & 55\% & 60\% & 62\% & 59\% & 36\% & 62\% & 63\% & 49\%\\
      \bottomrule
    \end{tabular}
    }
  \vspace{2mm}
  \caption{Comparing different templates $T_1$--$T_{15}$ and their impact on the F1-score in the zero-shot setting on CoNLL03. The results suggest that the performance of a template depends mainly on its naturalness.}
  \label{tab:template-results-CoNLL}    
\end{table*}

\begin{table*}[tbp]
  \centering
  \footnotesize
  \scalebox{0.9}{
    \centering
    \begin{tabular}{p{16mm}ccccccccccccccc}
      \toprule
      \textbf{} & \textbf{$T_1$} &\textbf{$T_2$}  & \textbf{$T_3$}
      & \textbf{$T_4$} &\textbf{$T_5$} &\textbf{$T_6$}  &\textbf{$T_7$} &\textbf{$T_8$} & \textbf{$T_9$}& \textbf{$T_{10}$} & \textbf{$T_{11}$}
      & \textbf{$T_{12}$}  & \textbf{$T_{13}$} & \textbf{$T_{14}$} & \textbf{$T_{15}$}\\
      \midrule
      LOC & 71\% & 72\% & 71\% & 64\% & 50\% & 69\% & 67\% & 72\% & 67\% & 67\% & 64\% & 68\% & 67\% & 68\% & 63\%\\
      
      PER & 49\% & 45\% & 43\% & 47\% & 24\% & 51\% & 51\% & 44\% & 48\% & 49\% & 51\% & 43\% & 45\% & 43\% & 47\%\\
      
      ORG & 47\% & 45\% & 44\% & 41\% & 36\% & 43\% & 42\% & 42\% & 46\% & 47\% & 47\% & 45\% & 44\% & 45\% & 42\%\\
      
      \midrule
      Micro-Avg. & 57\% & 55\% & 54\% & 52\% & 38\% & 56\% & 54\% & 54\% & 55\% & 56\% & 56\% & 54\% & 53\% & 54\% & 50\%\\
      \bottomrule
    \end{tabular}
    }
  \vspace{2mm}
  \caption{Impact of template choice in the zero-shot setting for OntoNotes}
  \label{tab:template-results-ontonotes}
\end{table*}

\vspace{-5mm}

\subsection{Domain Adaptation}
\label{sec:domain-adaptation}

In this section, we assess the extent to which our prompt-based approach can improve the performance of available baseline methods for NER in a domain adaptation setting. Specifically, we are interested in a setting where knowledge of the source domain should be transferred to a target domain for which the number of available training samples is very limited. This is of particular importance as annotating a new large corpus is often prohibitive. 

We consider three baselines: in the \textit{AGG} baseline, we merge the training data of the source and target domain and train a NER classifier on the resulting aggregated dataset. In the \textit{Fine-tuning} baseline, we first train the model on the source domain and then fine-tune it on the training set of the target domain. Both of these approaches have been shown to reach results competitive with other state-of-the-art methods \cite{metaNER}. In both cases, a BERT-large cased pre-trained LM followed by a linear layer is used as the NER classifier. 

The third baseline is \structshot, introduced by \cite{yang-katiyar-2020-simple}. It is based on a feature extractor module obtained by training a supervised NER model on the source domain. The generated contextual representations are used at inference time by a nearest neighbor token classifier. Additionally, label dependencies are captured by employing a viterbi decoder of a conditional random field layer and estimating the tag transitions using source domain.  

\begin{table}[t]
 \footnotesize
 \centering
 \scalebox{0.90}{
 \setlength{\tabcolsep}{6pt}
  \begin{tabular}{@{}lccccccc@{}}
    \toprule
    \textbf{Method} & \textbf{BC} &\textbf{BN}  & \textbf{MZ} & \textbf{NW} & \textbf{TC} & \textbf{WB}
    & \textbf{AVG}\\
    \midrule
    AGG & 46.3 & 47.9 & 46.9 & 52.7 & 51.7 & 43.8 & 48.2\\
    AGG+$T$ & 61.1 & 66.2 & 62.1 & 71.0 & 73.3 & 47.4 & 63.5 \\
    Fine-Tuning & 66.7 & 71.2 & 69.3 & 74.1 & 65.2 &  49.1 & 65.9 \\
    Fine-Tuning+$T$ & $\bm{72.0}$ & $72.1$ & 74.7 & $\bm{74.3}$ & 77.0 & 49.1 & 69.9  \\
    \structshot      & 63.2 & 70.6 & 71.6 & 71.8 & 67.3 & 51.2 & 65.9 \\ 
    \structshot +$T$ & 70.3 & $\bm{72.8}$ & $\bm{75.5}$ & 73.5 & $\bm{78.4}$ & $\bm{51.9}$ &  $\bm{70.4}$ \\
    \midrule
    In-domain & 91.6 & 94.3 & 94.1 & 93.2 & 76.9 & 67.1 & 86.2     \\ 
    \bottomrule
     \vspace{-2mm}
  \end{tabular}
  }
  \caption{Domain adaptation for NER: F1 score of using the OntoNotes and i2b2 datasets. Combining our prompt-based method $T$ with the Fine-Tuning approach achieves the best performance. 
   For all few-shot methods, we use K=100 samples of target domain training set. In contrast, the In-domain method uses all available training samples and serves as a topline. 
}
  \label{tab:ontonote_result}

\end{table}
\vspace{-4mm}

\begin{table*}[tbp]
 \footnotesize
 
 \begin{center}
 \setlength{\tabcolsep}{2pt}
 \scalebox{0.9}{
  \begin{tabular}{ccccccc}
    \toprule
     In-domain & AGG &AGG+$T$  & Fine-Tuning & Fine-Tuning+$T$ & \structshot  & \structshot +$T$ \\
    \midrule
   94.8 & 53.5 & 57.2 & 62.3 & $\bm{65.1}$ & 60.2  & 64.2\\
    \midrule
    \bottomrule
  \end{tabular}
  }
 \end{center}
 \vspace{-2mm}
  \caption{Results of domain adaptation for NER using OntoNotes as the source domain and i2b2 as the target domain.
}
  \label{tab:i2b2_result}
\end{table*}

\begin{table}[t]
 \footnotesize
 \resizebox{\textwidth}{!}{%
 \setlength{\tabcolsep}{2pt}
  \begin{tabular}{l|ccc|ccc}
    \toprule
    & \multicolumn{3}{c}{\textbf{one-shot}} & \multicolumn{3}{c}{\textbf{five-shot}} \\
    & Group A & Group B & Group C & Group A & Group B & Group C  \\
    \midrule
    Fine-Tuning & $7.4\pm2.4 $ &   $8.9 \pm 4.3 $  & $9.1 \pm 2.0 $  & $13.1 \pm 1.8$ &   $ 21.4 \pm 7.3$  & $ 20.8 \pm 4.2 $   \\
    Fine-Tuning+$T$ & $8.9\pm2.1 $ &  $11.8\pm 3.9 $  &  $13.6 \pm 1.7 $  & $14.9 \pm 1.6 $ &  $ 20.7 \pm 6.5 $  &  $22.5 \pm 2.6 $  \\
    \structshot &  $26.7\pm 4.8$ & $ 33.2\pm 15.6$ &  $23.0\pm11.4 $ &  $\bm{47.8}\pm 3.5$ & $56.5\pm9.3$ &  $53.4\pm3.2$ \\
    \structshot +$T$ &  $\bm{27.3}\pm 4.2$ & $\bm{33.6}\pm 14.1$ &  $\bm{26.5}\pm8.7 $ &  $46.0\pm 2.9$ & $\bm{57.7}\pm 7.8 $ &  $\bm{55.1}\pm 2.1 $ \\
    \bottomrule
     \end{tabular}
    }
    \vspace{2mm}
  \caption{Results of F1 scores on one-shot NER for different target labels. 
}
  \label{tab:group_one_shot}
\end{table}

\bigskip
%
%
Following \cite{wang-etal-2020-multi-domain-named}, we take one domain of the dataset as the target domain and the other domains as the source domain. We randomly select $K=100$ sample sentences from the target domain as our target training set. For selecting source and target labels, heterogeneous setup is adopted in which we choose \texttt{PERSON}, \texttt{ORG}, \texttt{PRODUCT} as source labels and \texttt{PERSON}, \texttt{ORG}, \texttt{PRODUCT}, \texttt{LOC}, \texttt{LANGUAGE}, \texttt{ORDINAL} as  target labels. This discrepancy between target and source labels makes transfer learning more challenging.

Table~\ref{tab:ontonote_result} depicts the results of our experiments on the various OntoNotes datasets averaged over five runs each. The table compares the performance of the baseline models with that of our hybrid approach. 
It also shows the results of an in-domain method, where we use all the training samples of target domain for training a fully supervised classifier. As is evident from this table, our prompt-based approach boosts the performance of all baseline models by a considerable margin. 
In our next experiment, we are interested in the impact of a greater discrepancy between source and target domain. We therefore take the OntoNotes~5.0 dataset as the source domain and the i2b2~2014 dataset
as the target domain.  
%

We use \texttt{PERSON}, \texttt{ORG}, \texttt{DATE}, \texttt{LOC} as source labels and \texttt{PERSON}, \texttt{ORG}, \texttt{DATE}, \texttt{LOC}, \texttt{PROFESSION} as target labels. Table~\ref{tab:i2b2_result} shows the results of our experiments. We observe the same pattern as before, i.e., combining our method with supervised baselines achieves the best performance. 
Lastly, we examine the role of tags on the performance of our method. To do so, we follow the same strategy as \cite{yang-katiyar-2020-simple} and split the entity categories of the OntoNotes into three non-overlapping groups:

\begin{itemize}
    \item  Group A: \texttt{ORG}, \texttt{NORP}, \texttt{ORDINAL}, \texttt{WORK OF ART}, \texttt{QUANTITY}, \texttt{LAW}
\item Group B: \texttt{GPE}, \texttt{CARDINAL}, \texttt{PERCENT}, \texttt{TIME}, \texttt{EVENT}, \texttt{LANGUAGE}
\item 
Group C: \texttt{PERSON}, \texttt{DATE}, \texttt{MONEY}, \texttt{LOC}, \texttt{FAC}, \texttt{PRODUCT}
\end{itemize}

When we pick a group, \eg A, as the target group, the corresponding tags are replaced with the \texttt{O}-tag in the training data. Thus, the model is trained only on source groups B and C. At inference time, we evaluate the model on the test set which contains only target labels (A). 
The results of this experiments are shown in Tables~\ref{tab:group_one_shot} 
for the one-shot and $5$-shot setting. We again observe the improvement of baseline models by integrating them into template approach. The amount of performance gain depends on the target group, of course. In particular, we get a large amount of improvement for group C of tags around $3.5$ in one-shot and $1.7$ in the five-shot setting.



\section{Conclusion and Future Work}
\label{sec:conclusion}

We proposed a novel, lightweight approach to NER for zero- and few-shot settings, using pre-trained language models to fill in cloze-style prompts. It is based on extracting information available in PLMs and utilizes it to labels named entity instances identified by a domain-independent POS tagger. Results show that masked language models have a better performance in this setting compared with auto-regressive language models.
We explored a wide range of possible prompts with different datasets. We observed that the proposed method is robust against contextual details of prompts. This is of practical significance in the low-resource setting where there is not enough data to tune the model. 
Our method is simple, general and can be used to boost the performance of available domain adaptation baselines. We also propose a hybrid approach that can easily combine the template method with any other supervised/unsupervised classifier, and demonstrated the effectiveness of this hybrid approach empirically. 

Further work could investigate the possibility of fine-tuning templates while having access only to a few training samples. It would be also interesting to explore more sophisticated approaches for combining the predictions of the template model and other few-shot NER baselines. 
Two aspects of our approach currently require manual intervention, the template and the representative word lists. Finding ways to determine these fully automatically is another interesting direction to explore. As mentioned before, one way to extract representative words is by making use of word embeddings like GloVe. Indeed, we found that almost all subsets of our representative words perform fairly well in practice. We leave automatically extraction of representative words and its evaluation to future work. 

\section{Acknowledgments}
\vspace{-2mm}
This work was funded by the EU-funded Horizon 2020 projects: COMPRISE (\texttt{http://www.compriseh2020.eu/}) under grant agreement No. 3081705 and ROXANNE under grant number 833635. 
%
%
\bibliographystyle{splncs04}
\bibliography{mybibliography}

\end{document}